\documentclass[a4paper,fleqn]{cas-sc}

\usepackage[numbers]{natbib}
\usepackage{subcaption}
\usepackage{float}
\usepackage{placeins}

\def\tsc#1{\csdef{#1}{\textsc{\lowercase{#1}}\xspace}}
\tsc{WGM}
\tsc{QE}

\begin{document}
\let\WriteBookmarks\relax
\def\floatpagepagefraction{1}
\def\textpagefraction{.001}

\shorttitle{}    

\shortauthors{}  

\title [mode = title]{When AI and Experts Agree on Error: Intrinsic Ambiguity in Dermatoscopic Images}  


\tnotetext[1]{} 

%

\author[1]{Loris Cino}





\credit{Conceptualization, Methodology, Formal analysis, Writing – original draft}

\affiliation[1]{organization={Sapienza - Dipartimento di Ingegneria informatica, automatica e gestionale (DIAG)
},
            addressline={V. Ariosto, 25}, 
            city={Rome},
            postcode={00185}, 
            state={Lazio},
            country={Italy}}

\author[2]{Pier Luigi Mazzeo}
\credit{Methodology, Formal analysis, Writing – review \& editing}




\credit{Conceptualization, Methodology, Writing – original draft}

\affiliation[2]{organization={
Istituto di Scienze Applicate e Sistemi Intelligenti (ISASI), Consiglio Nazionale delle Ricerche (CNR)},
            addressline={Via Monteroni s.n}, 
            city={Lecce},
            postcode={73100}, 
            country={Italy}}

\author[3]{Alessandro Martella}

\credit{Supervision, Project administration, Resources}

\affiliation[3]{organization={Dermatologia Myskin, Poliambulatorio Specialistico Medico-Chirurgico},
            addressline={ Via S. Marco, 21}, 
            city={Tiggiano},
            postcode={73030}, 
            state={Puglia},
            country={Italy}}
            
\author[4]{Giulia Radi}

\credit{Resources}

\affiliation[4]{organization={
            AST Pesaro-Urbino},
            addressline={Via Borsellino 4}, 
            city={Fano},
            postcode={ 60019 }, 
            state={Marche},
            country={Italy}}
            
\author[5]{Renato Rossi}

\credit{Resources}

\affiliation[5]{organization={
            La Rocca Skin Medical Center},
            addressline={ Via Marchetti, 110}, 
            city={Senigallia},
            postcode={ 61122 }, 
            state={Marche},
            country={Italy}}
            
\author[2]{Cosimo Distante}
\cormark[1]
\ead{cosimo.distante@cnr.it}
\credit{Supervision, Project administration, Funding acquisition}

\cortext[1]{Corresponding author}


\begin{abstract}
The integration of artificial intelligence (AI), particularly Convolutional Neural Networks (CNNs), into dermatological diagnosis demonstrates substantial clinical potential. While existing literature predominantly benchmarks algorithmic performance against human experts, our study adopts a novel perspective by investigating the intrinsic complexity of dermatoscopic images 
Through rigorous experimentation with multiple CNN architectures, we isolated a subset of images systematically misclassified across all models—a phenomenon statistically proven to exceed random chance. To determine if these failures stem from algorithmic biases or inherent visual ambiguity, expert dermatologists independently evaluated these challenging cases alongside a control group. The results revealed a collapse in human diagnostic performance on the AI-misclassified images. First, agreement with ground-truth labels plummeted, with Cohen’s kappa dropping to a mere 0.08 for the difficult images, compared to a 0.61 for the control group. Second, we observed a severe deterioration in expert consensus; inter-rater reliability among physicians fell from moderate concordance (Fleiss’ kappa = 0.456) on control images to only modest agreement (Fleiss’ kappa = 0.275) on difficult cases. We identified image quality as a primary driver of these dual systematic failures. To promote transparency and reproducibility, all data, code, and trained models have been made publicly available.
\end{abstract}


\begin{highlights}
\item Multiple AI algorithms systematically misclassify the same images.
\item Human diagnostic accuracy and consensus degrade on these AI-misclassified cases.
\item Image artifacts and low quality partially explain these shared diagnostic errors.
\end{highlights}


\maketitle

\section{Introduction}
\label{sec:introduction}
Skin tumors, including Basal Cell Carcinoma (BCC) \cite{BCC}, Squamous Cell Carcinoma (SCC), and Melanoma (MEL), represent some of the most common and clinically significant dermatological malignancies worldwide. In light of the significant health risks they pose, dermatoscopy has emerged as a primary non-invasive technique for their early detection. This method is widely adopted in clinical practice, as it enhances the visualization of subsurface skin structures, supports more accurate diagnosis, and ultimately contributes to earlier intervention and reduced mortality.

Despite dermatoscopy effectiveness, histopathological examination remains the gold standard for diagnosis, as the dermatoscopic characteristics assessed using the ABCDE (Asymmetry, Border, Color, Diameter, Evolution) rule \cite{ABCDE} may not be decisive for a specific cutaneous neoplasm, especially in the case of difficult lesions that appear strikingly similar. In addition to its role in oncologic dermatology, dermatoscopy is extensively utilized in the evaluation of chronic inflammatory skin conditions, which affect approximately 25–30\% of the population, including psoriasis. Furthermore, it is employed in the assessment of parasitic infestations and the examination of skin appendages, such as nails and hair \cite{inflammatory}. 

As dermatoscopy is a fundamentally image-centric field, the application of deep neural networks—driven by the remarkable advancements in CNNs—represents a natural evolution in diagnostic analysis \cite{survey_medical_imageing}. 
Over the years, various methodologies have been explored to integrate AI into dermatological diagnostics, encompassing both deep learning-based and alternative approaches. A seminal study demonstrated that CNN-based models could achieve diagnostic performance comparable to that of expert dermatologists, marking a significant step in AI-driven dermatology \cite{esteva}. 

Other works demonstrated that performance can be enhanced by utilizing multiple models and aggregating their predictions to generate a more robust and reliable outcome \cite{Harangi2018, Gannour2023, NancyJane2023}, a methodology commonly referred to as ensemble learning. Another strategy for improving model efficacy involves incorporating additional patient-specific information, such as age and the anatomical location of the lesion \cite{Hhn2021}. Notably, Gannour et al. \cite{Gannour2023} integrate both ensemble methods and individual patient data in their analysis. However, the extent to which these approaches—ensemble learning and patient-specific data—contribute to overall model improvement remains an open question. Furthermore, to the best of the authors' knowledge, no prior work has employed a meta-classifier, a model fitted on other models predictions, CNNs in this case.



Despite the wide literature deep learning is not widely adopted in the dermatoscopy \cite{Wei2024, Krakowski2024}, an empirical evidence is that the United Stated Food and Drug Administration approved several AI algorithms for medical use but none of them regard dermatology \cite{Wu2021}. 
The limited adoption of AI-based dermatological diagnostic tools in clinical practice can be attributed to several technical factors: (i) performance degradation when these algorithms are applied to OOD; (ii) issues related to data quality \cite{Wei2024}; and (iii) the unclear impact of AI-assisted diagnosis on clinical decision-making. While all these factors are essential for a comprehensive understanding of the literature, this study focuses primarily on the last two aspects. Specifically, we analyze the diagnostic accuracy of expert dermatologists and AI models at the image level, demonstrating that both human experts and AI systems struggle with the same subset of images. Furthermore, we identify a significant presence of low-quality images within commonly used dermatological datasets and propose an automated approach for detecting blurred images.

\subsection{Out-of-distribution performance}
The investigation into the out-of-distribution performance of machine learning models applied in dermatology has garnered considerable attention in contemporary research \cite{Li2023}. Recent studies have highlighted a decline in model performance when evaluated on images that deviate from those included in the training dataset \cite{Wei2024}. This decline is particularly pronounced when models are tested on individuals from ethnicities and with skin colors that are underrepresented in the training data, highlighting the need to test the model on different Fitzpatrick skin type labels \cite{groh2021evaluating, groh2022towards}. 
Interestingly, this phenomenon is also observed among physicians, who tend to achieve higher diagnostic accuracy on skin types with which they are most familiar \cite{Wei2024}.


\subsection{Data Quality}
A key challenge in AI-based dermatological research is the heterogeneity of available datasets, given the diversity of skin diseases and imaging modalities. This study focuses on leveraging dermatoscopic images, which are typically more rare, for skin cancer identification. The most known datasets are HAM10000 \cite{ham10000_2018}, ISIC \cite{bcn20000_2019, skin_2018}, and PH2 \cite{ph2} for skin disease classification \cite{Khan2021, Cao2023}. Additionally, recently introduced datasets, including crowdsourced repositories \cite{google_dataset} and densely annotated datasets such as SkinCon \cite{skincon}, further expand the resources available for model development but they don't contain dermatoscopic images.

While the increasing availability of dermatological images addresses the issue of data quantity, data quality is often neglected. This remains a critical oversight, as high-quality data is essential in machine learning, particularly for high-stakes applications like healthcare \cite{Jones2022}. Considering the dermatoscopic images there are several factor that can impact the quality of the image: focus, angle, lightning, color representation \cite{Winkler2021, Maier2022}. Also image size plays a pivotal role in the performance of AI algorithms in dermatology \cite{Cino2022}. The image quality had gained a so critical relevance that Daneshjou et al. \cite{Daneshjou2022} presented some guidelines for dermatoscopic images. During the manual evaluation, dermatologists identified a subset of low-quality images, the majority of which were characterized by being out-of-focus. Motivated by these observations, we employed standard algorithms for blur detection to help to systematically identify additional blurred images within the dataset.

\subsection{Dermatologist and AI}
The encouraging outcomes of machine learning in dermatology have stimulated numerous comparative studies assessing algorithmic performance against dermatologists of varying experience levels \cite{expert_german}. \citet{survey_AIvsCNNs} provided an in-depth analysis of these evaluations, reviewing 19 studies on CNNs in skin disease imagery, 11 of which focused specifically on dermatoscopy \cite{Brinker2019, Haenssle2020}. A consistent observation across this literature is that CNNs frequently match or surpass the diagnostic capabilities of dermatologists in tasks ranging from classification to therapeutic decision-making.

\begin{figure}[]
    \centering
    \includegraphics[width=.8\linewidth]{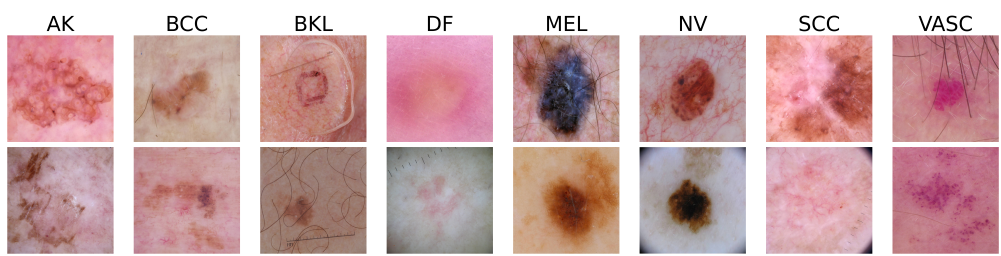}
    \caption{
    Representative dermatoscopic images for each diagnostic category included in the ISIC 2019 dataset. The first row shows examples correctly classified by all neural network models, while the second row presents images misclassified by the models. Each column corresponds to one diagnostic class (AK, BCC, BKL, DF, MEL, NV, SCC, VASC).
    }
    \label{fig:examples}
\end{figure}

Recent literature has shifted beyond mere performance comparisons to explore the interaction and integration of experts and algorithms. \citet{Zakhem2021} highlight that while dermatologist expertise is crucial for data collection, identifying biases, and technological application, clinicians remain underrepresented in the publications describing these tools. Furthermore, research by \citet{Heinlein2024} emphasizes the impact of AI on clinical diagnosis, framing deep learning as a supportive \textit{assistive tool}. For instance, AI can provide critical decision support when clinicians encounter skin phototypes that are uncommon in their specific geographic region \cite{Minagawa2020}. Most recently, \citet{Yamamura2025} demonstrated that large language models (LLMs) can achieve diagnostic accuracy comparable to dermatologists and provide valuable assistance in complex cases.

Despite this growing interest, the real-world clinical impact of AI remains nuanced. While many studies highlight beneficial outcomes \cite{Tschandl2020, Krakowski2024}, others warn of a negative impact when an algorithm misclassifies an image \cite{Han2022}. There is, however, a general consensus that AI provides the greatest benefit to less experienced practitioners \cite{Krakowski2024}. 

Moving beyond traditional comparative analysis, we examine the integration of expert and AI insights through the lens of shared error patterns. Unlike previous studies, we shift the focus toward the intrinsic difficulty of individual images rather than a simple human-versus-machine competition. The understanding of these images could improve the machine learning models while provide insight on how and when use them in clinical practice.

The principal findings of this work are summarized as follows:
\begin{enumerate}
    \item Images misclassified by CNNs also exhibit a higher rate of error among dermatologists.
    \item Expert evaluation revealed that a subset of \textit{difficult images} were actually of inadequate quality, leading us to develop an automated analysis to identify and mitigate the presence of blurred images.
\end{enumerate}

Finally, acknowledging the lack of reproducibility in prior studies \cite{Wei2024}, we provide full access to our code and model weights via our GitHub repository to support transparent and reproducible research in this domain. \footnote{https://github.com/Loris697/skin-disease-expert}.

\section{Materials and Methods}
\label{sec:methods}
The study utilizes five pretrained CNNs previously described detailed in \cite{Cino2025}, specifically ResNeXt-50, ResNet-152, EfficientNet-B4, EfficientNet-B5, and EfficientNet-B6. The primary objective is to systematically analyze images that are consistently misclassified by all neural networks. To gain deeper insights into the characteristics of these images, we also present them to expert dermatologists for evaluation. 

\subsection{Training Protocol}
All models were initially pretrained on the ImageNet dataset \cite{IMAGENET}
followed by fine-tuning on the ISIC 2019 dataset \cite{ham10000_2018, bcn20000_2019, skin_2018}. ISIC images are used without further color transformation. For illustrative purposes, representative images from each diagnostic class in the dataset are shown in Figure~\ref{fig:examples}. 


We employed a 5-fold cross-validation approach, this methodology ensured that each image was included in the validation set at least once. 
In total, 25 models were trained (five architectures across five data splits), requiring approximately two weeks of training on a Titan RTX GPU with 24GB RAM. 


\subsection{Analysis of Common Misclassifications}

Based on the models' predictions, we observed that the number of images consistently misclassified by all networks—hereafter referred to as \textit{difficult} images—is significantly higher than what would be expected by chance. Specifically, we tested the hypothesis that these errors arise from a null model, where misclassifications are independent of image identity and no intrinsic difficulty exists.

To statistically evaluate this hypothesis we performed a stratified 
permutation test \cite{Holt2023}, a non-parametric approach does not 
rely on distributional assumptions and, importantly, accounts for the 
lack of independence among models arising from their training on the same dataset.
In our setting, each model constitutes a stratum: for model $m$, we permute its 
error vector across images while keeping the total number of errors made by that 
model fixed. This procedure breaks the association between a specific image and 
the errors of each model, while preserving the overall error rate of each 
classifier.

Formally, our null hypothesis states that:
\begin{align}
H_0 &: \text{All images have the same probability of being misclassified;}
\end{align}

Under this null, permuting each model's error profile across images should not 
affect the distribution of the number of images misclassified by all models. 
We compute the permutation distribution of this statistic using a Monte Carlo 
approximation, as exact enumeration is infeasible for our dataset size.

We performed 100,000 random shuffles of the
models' error patterns and computed 
the $p$-value as the proportion of permutations that resulted in a statistic 
more extreme than, or equal to, the one observed in the empirical data. 
We adopted a significance level of 5\%.

The rejection of the null hypothesis would indicate that certain images are consistently misclassified across all models more frequently than expected by chance. This pattern suggests either correlated model predictions or intrinsic characteristics that make these images inherently difficult to categorize. To investigate these \textit{difficult} cases, we tasked six expert dermatologists with independently evaluating the subset of images most frequently misclassified. This human assessment aimed to determine whether these instances also pose challenges for specialists. For comparison, the evaluation included a control group of the evaluation included a control group of randomly selected images that were correctly classified by all models.

\subsection{Dermatologists’ Experience}
Each expert dermatologist that evaluated the image has more than 20 years of experience in the field.
These specialists are highly experienced dermatologists with advanced expertise in dermoscopy.
Over the years, they have contributed significantly to the field through clinical practice,
scientific research, and peer-reviewed publications. Their work includes studies on magnified
dermoscopy, diagnostic accuracy, and innovative imaging techniques.

\subsection{Collection of Diagnoses}
The identified images were uploaded to a web-based platform developed using WordPress and Bootstrap for dermatological assessment, to maintain an unbiased evaluation, ground-truth labels were not revealed to experts. The portal is accessible at \underline{http://150.146.211.50/}, with access credentials available upon request from the authors. More informations about the portal can be found in Appendix \ref{a:website}.

To facilitate detailed visual inspection, the interface displayed relevant patient data and allowed dermatologists to enlarge the provided images. For the diagnostic phase, the system offered a predefined list of conditions, an 'Other' option for uncertain cases, and a comment field for additional remarks.

A total of 270 images were uploaded, divided into two groups: (a) 190 \textit{challenging} images and (b) a control group of 80 images correctly classified by all networks. While we aimed for a balanced distribution of at least 10 images per class, this was not always possible due to the scarcity of certain misclassifications. Specifically, the dermatofibroma (DF) and actinic keratosis (AK) classes included only five and three 'difficult' images, respectively. The first row of Figure \ref{fig:examples} displays example images from the control group (b), while the second row shows examples from the challenging group (a)





\subsection{Analysis of the Diagnosis}
The diagnostic outcomes provided by dermatologists, collected using the aforementioned website, were systematically analyzed across various groups, among different dermatologists, and compared to the predictions made by neural networks and ground truth as well. This comparison was quantitatively assessed by using Cohen's kappa \cite{Cohen1960} to measure the agreement. Specifically, we compared the diagnostic results obtained by human experts on the "difficult" images, the control group, and the findings reported in other works within the literature, demonstrating that the same images consistently present challenges for dermatologists as well. In order to ensure the most objective diagnostic labels, we included only cases for which a majority agreement among dermatologists was reached. Because the number of dermatologists was even, the most experienced clinician performed an additional evaluation, ensuring an odd number of total assessments and preventing many ties in cases lacking clear consensus. Diagnoses indicating uncertainty, submitted via the \textit{Other} option, were excluded from the analysis.

Furthermore, contingency tables were employed to uncover additional correlations between diagnoses \cite{everitt1992}. The utility of contingency tables becomes particularly evident when certain types of diseases are frequently misclassified as one another. For instance, melanoma and nevus, which are distinct conditions, are commonly misidentified as each other. This methodological approach enables the identification of diagnostic agreement patterns and discrepancies, thereby enhancing the understanding of both human and neural network predictive accuracy.

To assess the inter-rater reliability among the six expert dermatologists, we calculated Fleiss' kappa \cite{fleiss1971}, a statistical measure appropriate for evaluating agreement among more than two independent raters assigning categorical ratings.

\subsection{Low quality Images}
In the process of analyzing image diagnoses, it was noted that among the 270 images uploaded for review, dermatologists faced challenges in diagnosing 18 images due to their blurred or low-quality nature. High-resolution images are crucial because they provide detailed visualization of morphological features essential for accurate skin lesion classification. Such clarity allows dermatologists to discern subtle distinctions between benign and malignant lesions, thereby reducing diagnostic errors \cite{Ferrara2020}. Moreover, the performance of CNNs is significantly influenced by the quality of the input images, even if a small fraction of the dataset has low quality \cite{Pei2021}. Low-quality, blurred images often lack the necessary detail, resulting in reduced feature extraction capability by the CNN. 

To systematically identify additional blurred images within the dataset, three distinct techniques were employed: the Laplacian filter, the Fourier transform, and wavelet transformation. Each method was utilized to isolate the 50 images with indices indicative of potential blurriness. 

The Laplacian filter, a second-order derivative filter, plays a pivotal role in identifying areas within an image that exhibit rapid intensity changes—commonly referred to as edges \cite{XinWang2007}. This filter computes the second derivatives of the image's intensity in both the horizontal and vertical directions, unify these measurements into a unified metric. 

Furthermore, the Fourier transform is fundamental in converting a signal from the spatial to the frequency domain. This transformation facilitates the decomposition of an image into its constituent sinusoidal components, thereby segregating the diverse frequency elements present within. Blurring effects are known to attenuate the high-frequency components of an image \cite{Narwaria2012}. 

The third technique employed, the wavelet transform, serves as a mathematical tool designed to decompose a signal into its integral components across various scales.
Similar to the previously discussed methods, the application of wavelet transform in detecting blur within images hinges on the principle that blurring predominantly attenuates the high-frequency components of an image \cite{HanghangTong}. 

The comprehensive methodological framework outlined above provides a robust foundation for analyzing the subset of “difficult” images. It also enhances the interpretability of the results and supports a clearer understanding of the overall findings of the study.


\section{Results}
\label{sec:results}
The evaluation of our methodology focuses on three key aspects: (1) the statistical significance of systematic network errors; (2) the alignment among algorithmic predictions, ground-truth labels, and human expert diagnoses; and (3) the impact of low-quality images. Specifically, Section \ref{subsec:common_misclassification} reports the outcomes of the permutation test used to evaluate the expected number of simultaneous misclassifications across all CNNs. Section \ref{subsec:result_diagnosis} presents an analysis of the alignment between the dermatologists' diagnoses and the dataset labels across different image groups. Here, we compare the diagnostic accuracy on \textit{difficult images} against the control group, and contextualize our findings within the existing literature. Finally, Section \ref{subsec:result_low_quality} details the findings related to low-quality and duplicated images identified within the datasets. The catalog of low-quality images compiled during our study provides a valuable resource for future research aimed at enhancing overall dataset quality.

\subsection{Common Misclassification}
\label{subsec:common_misclassification}



Firstly, we formally assess whether the observed proportion of misclassified images deviates significantly from random chance, we conduct a stratified permutation test, calculating the p-values using Monte Carlo methods. Under the null hypothesis, the probability of a given image being misclassified is equal for each image.









\begin{figure}[]
    \centering
    \includegraphics[width=.7\linewidth]{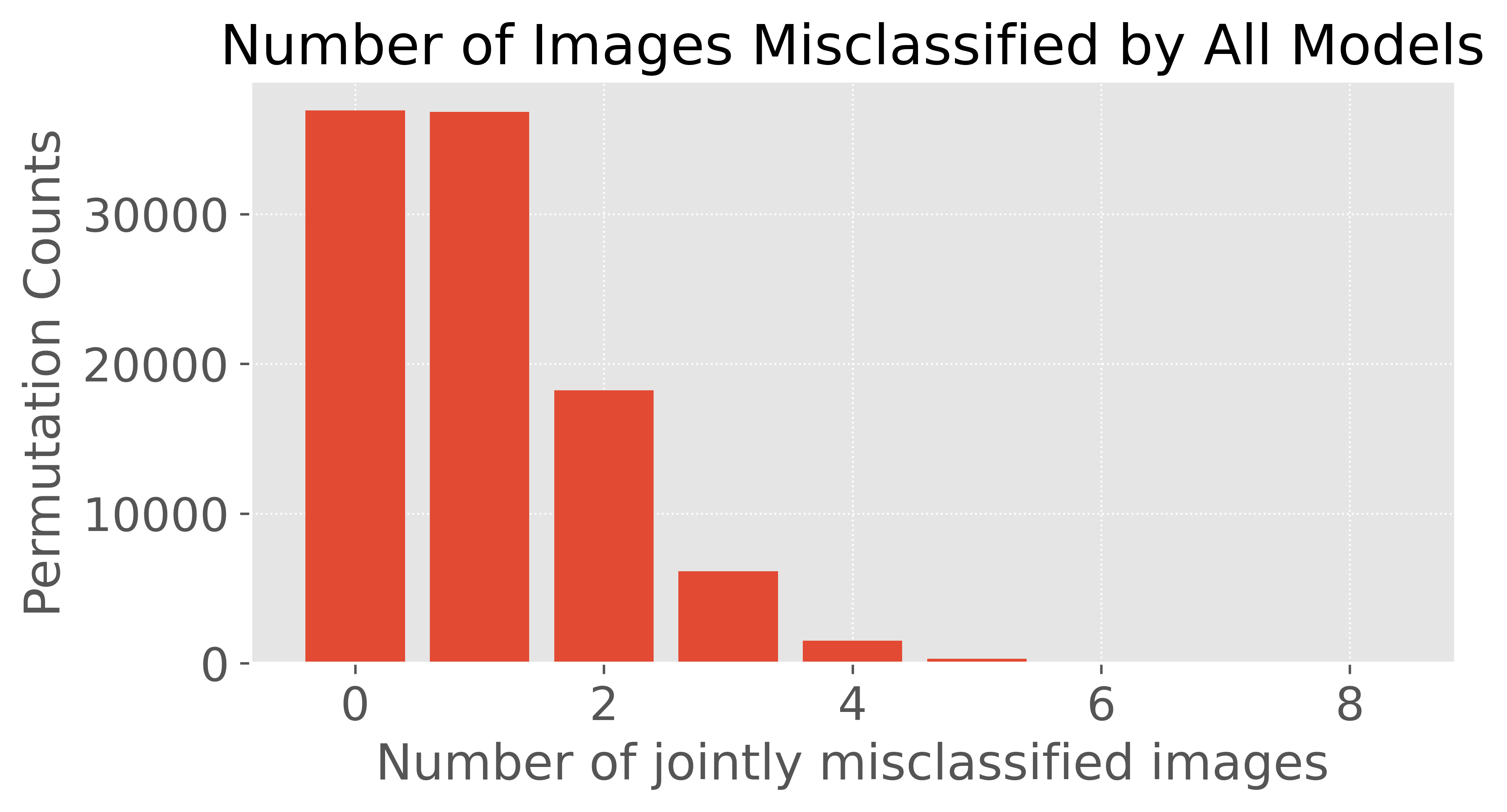}
    \caption{
    Distribution of the number of images jointly misclassified by all models, obtained through a stratified permutation test used to assess whether the observed level of simultaneous mistakes exceeds what would be expected under the null hypothesis.
    }
    \label{fig:permutation}
\end{figure}

Figure \ref{fig:permutation} depicts the distribution of the number of images consistently misclassified by all models across 100,000 iterations of the stratified permutation test. 
The distribution of the number of images misclassified by all models is tightly concentrated, with a 95\% confidence interval ranging approximately from 0 to 3. These results stand in stark contrast to the 823 images consistently misclassified in the empirical experiment.

The resulting p-value is effectively zero: none of the $100{,}000$ permutations produced a number of jointly misclassified images comparable to the empirical observation,
we therefore reject the null hypothesis and conclude that the substantial number of consistently misclassified images cannot be attributed to random chance.

This finding suggests the presence of systematic patterns underlying these errors, potentially arising from intrinsic image characteristics, biases introduced during model training, or dependencies among model predictions due to shared training data and architectures. In the subsequent experiments, we investigate whether this phenomenon stems from image-specific attributes or network-related factors by introducing an independent set of predictions: the clinical assessments provided by expert dermatologists for these challenging cases.


\subsection{Analysis of the Diagnosis}
\label{subsec:result_diagnosis}
After the collection the diagnosis of human experts are compared between the \textit{difficult group} (a) and the control group (b).  We verify the source of each label whenever available. We were able to reconstruct the label source only for the HAM10000 dataset. Among the \textit{difficult} images, only 254 originate from this dataset, and of these, 245 were confirmed through histological examination, the gold standard for diagnosing skin cancer. Other datasets report a high percentage of histologically validated images. This gives us confidence that the low diagnostic metrics reported are likely not due to labeling errors.

Table \ref{tab:perclass} details the diagnostic performance of the dermatologists across the two image subsets. To establish a baseline, we first evaluated the consensus diagnoses for images correctly classified by all neural networks. In this control group, the human experts achieved a sensitivity of 0.90 and a specificity of 0.875 for melanoma, along with a sensitivity of 0.625 and a specificity of 0.98 for BCC. These metrics align well with, and in some aspects exceed, general clinical performance reported in the literature; for instance, \citet{Dinnes2018a} reported a MEL sensitivity of 0.81, and a BCC sensitivity of 0.93 \cite{Dinnes2018b}, both at a fixed specificity of 0.80. This suggests that images easily classified by AI algorithms are also reliably diagnosed by human experts.

In contrast, when evaluating the \textit{difficult images}, human diagnostic accuracy experienced a substantial decline, falling significantly below the literature baselines. For the MEL class, sensitivity plummeted to 0.53 and specificity to 0.65. Most notably, the human experts failed to correctly identify a single BCC case within this challenging subset. This disparity underscores that images posing classification challenges to the CNNs are similarly challenging for dermatologists, affirming the complexity and inherent difficulty of accurately diagnosing these particular cases. This insight not only validates the performance of the CNNs but also highlights potential areas for enhancing diagnostic accuracy both in automated systems and clinical practice. Interestingly, both CNNs and physicians consistently excelled at identifying VASC, suggesting that visual inspection is generally sufficient for this disease class.

\begin{table}[]
    \centering
    \caption{
    Per-class sensitivity and specificity of the dermatologists’ diagnoses for the two image groups: the easy group (images correctly classified by all models) and the difficult group (images consistently misclassified by the CNNs).
    }
    \label{tab:perclass}
    \resizebox{0.5\textwidth}{!}{
    \begin{tabular}{l|cc|cc}
        \toprule
        & \multicolumn{2}{c|}{\textbf{Easy}} & \multicolumn{2}{c}{\textbf{Difficult}} \\
        \cmidrule(lr){2-3} \cmidrule(lr){4-5}
        \textbf{Class} & \textbf{Sensitivity} & \textbf{Specificity} & \textbf{Sensitivity} & \textbf{Specificity} \\
        \midrule
        AK   & 0.3 & 0.9531 & 0.125 & 0.8529 \\
        BCC  & 0.625 & 0.9848 & 0.0000 & 0.9565 \\
        BKL  & 0.5 & 0.9687 & 0.0454 & 0.9846 \\
        DF   & 0.625 & 1.0000 & 0.0000 & 1.0000 \\
        MEL  & 0.9 & 0.875 & 0.5344 & 0.6489 \\
        NV   & 0.7 & 0.9531 & 0.2692 & 0.7619 \\
        SCC  & 0.6250 & 0.8939 & 0.1818 & 0.9148 \\
        VASC & 1.0 & 0.9843 & 0.6666 & 0.9731 \\
        \bottomrule
    \end{tabular}
    }
\end{table}

 


To quantify overall diagnostic accuracy, we computed the proportion of dermatologist assessments that matched the ground-truth labels. For the \textit{difficult} group, the accuracy was merely 29.6\%, with a corresponding Cohen’s kappa of 0.08, indicating negligible agreement beyond chance. In stark contrast, diagnostic accuracy for the control group was 66.2\%, with a Cohen’s kappa rising to 0.61.

These findings provide additional context to the observations by \citet{Han2022}. This study demonstrated that physicians exhibit lower diagnostic accuracy when assessing images that are misclassified by AI models. This phenomenon, however, may not necessarily be attributable to the influence of AI algorithms. Instead, it is plausible that the observed reduction in diagnostic performance is primarily due to the intrinsic difficulty of the images themselves.

The results of this comparison are detailed in the contingency tables referenced as Table \ref{table_consensus_complicated} and Table \ref{table_consensus_easy}. 
Focusing exclusively on images misclassified by all neural networks, we observe that several images were diagnosed by human experts as belonging to classes that differ significantly in visual characteristics from their actual labels; for instance, some images labeled as MEL were predominantly diagnosed by dermatologists as AK. In contrast, considering only images correctly classified by all neural networks, wrong diagnosis occur mainly between visually similar classes, such as NV and MEL.


Furthermore, the sensitivity in images in malignant is usually higher showing that the physicians tend to diagnosis more frequently this kind of disease, this is reasonable considering that not correctly diagnose a MEL is much more dangerous, works that compare human expert with AI diagnois should take this into account. 


\begin{table}[]
    \centering
    \caption{Contingency tables reporting the distribution of true lesion labels (rows) versus the diagnoses assigned by the majority of dermatologists (columns).}
    \label{table_consensus_combined}

    \begin{subtable}[t]{0.48\textwidth}
        \centering
        \caption{Images misclassified by all neural network models. The “DF” column is absent as no image received a majority diagnosis of DF.}
        \label{table_consensus_complicated}
        \resizebox{\linewidth}{!}{%
        \begin{tabular}{c|c|c|c|c|c|c|c}
        \hline
        \textbf{Diag} & \textbf{AK} & \textbf{BCC} & \textbf{BKL} & \textbf{MEL} & \textbf{NV} & \textbf{SCC} & \textbf{VASC} \\
        \textbf{Label}  &  &  &  &  &  &  & \\
        \hline
        \textbf{AK} & 2 & 0 & 0 & 7 & 2 & 5 & 0 \\
        \textbf{BCC} & 6 & 0 & 0 & 0 & 2 & 5 & 1 \\
        \textbf{BKL} & 3 & 1 & 1 & 10 & 6 & 0 & 1 \\
        \textbf{DF} & 0 & 0 & 0 & 0 & 1 & 0 & 1 \\
        \textbf{MEL} & 5 & 0 & 2 & 31 & 18 & 1 & 1 \\
        \textbf{NV} & 2 & 3 & 0 & 13 & 7 & 1 & 0 \\
        \textbf{SCC} & 4 & 2 & 0 & 3 & 0 & 2 & 0 \\
        \textbf{VASC} & 0 & 0 & 0 & 0 & 1 & 0 & 2 \\
        \hline
        \end{tabular}%
        }
    \end{subtable}\hfill 
    \begin{subtable}[t]{0.48\textwidth}
        \centering
        \caption{Group of comprehending images correctly classified by all neural network models. In this case all the diagnosis and ground-truth are present. }
        \label{table_consensus_easy}
        \resizebox{\linewidth}{!}{%
        \begin{tabular}{c|c|c|c|c|c|c|c|c}
        \hline
        \textbf{Diag} & \textbf{AK} & \textbf{BCC} & \textbf{BKL} & \textbf{DF} & \textbf{MEL} & \textbf{NV} & \textbf{SCC} & \textbf{VASC} \\
        \textbf{Label}  &  &  &  &  & & &  & \\
        \hline
        \textbf{AK} & 3 & 0 & 1 & 0 & 0 & 1 & 4 & 1 \\
        \textbf{BCC} & 0 & 5 & 0 & 0 & 0 & 1 & 2 & 0 \\
        \textbf{BKL} & 2 & 0 & 5 & 0 & 1 & 1 & 1 & 0 \\
        \textbf{DF} & 0 & 0 & 1 & 5 & 2 & 0 & 0 & 0 \\
        \textbf{MEL} & 1 & 0 & 0 & 0 & 9 & 0 & 0 & 0 \\
        \textbf{NV} & 0 & 0 & 0 & 0 & 3 & 7 & 0 & 0 \\
        \textbf{SCC} & 0 & 1 & 0 & 0 & 2 & 0 & 5 & 0 \\
        \textbf{VASC} & 0 & 0 & 0 & 0 & 0 & 0 & 0 & 10 \\
        \hline
        \end{tabular}%
        }
    \end{subtable}

\end{table}

Beyond a decline in diagnostic validity, we also observed a marked reduction in inter-rater reliability when comparing the diagnosis of different physicians. Specifically, the Fleiss’ kappa among dermatologists was 0.275 for the difficult images—indicating only modest agreement—whereas for the control group it increased to 0.456, corresponding to a moderate level of concordance.

\subsection{Low quality images}
\label{subsec:result_low_quality}
During the course of the experiments, a subset of images was flagged by the doctors as being of low quality. Such images, when included in the training set, have the potential to introduce noise and degrade the model's performance, or, if present in the test set, they may hinder proper evaluation. 

\begin{figure}[]
    \centering
    
    \begin{subfigure}{0.23\textwidth}
        \centering
        \includegraphics[width=\linewidth]{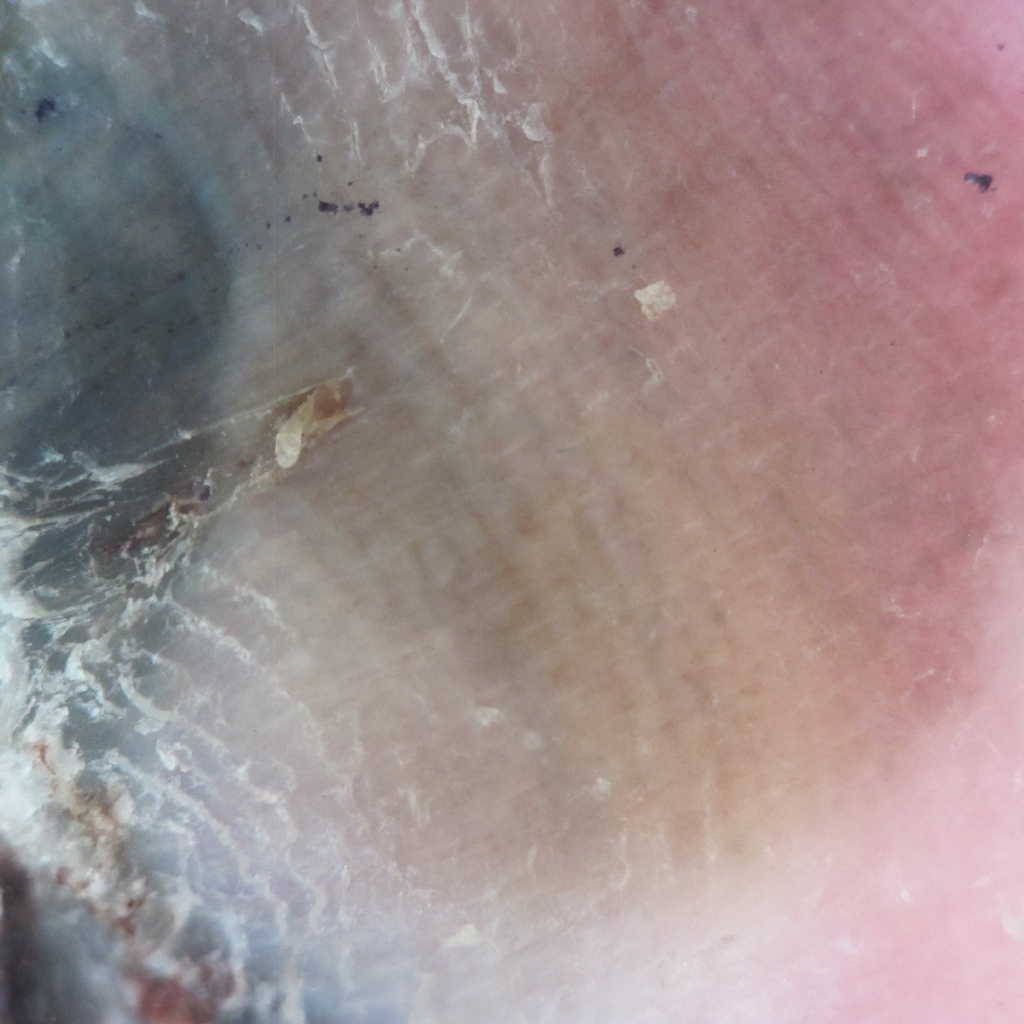}
        \caption{ISIC\_0053699}
        \label{fig:low_med_left}
    \end{subfigure}\hfill
    \begin{subfigure}{0.23\textwidth}
        \centering
        \includegraphics[width=\linewidth]{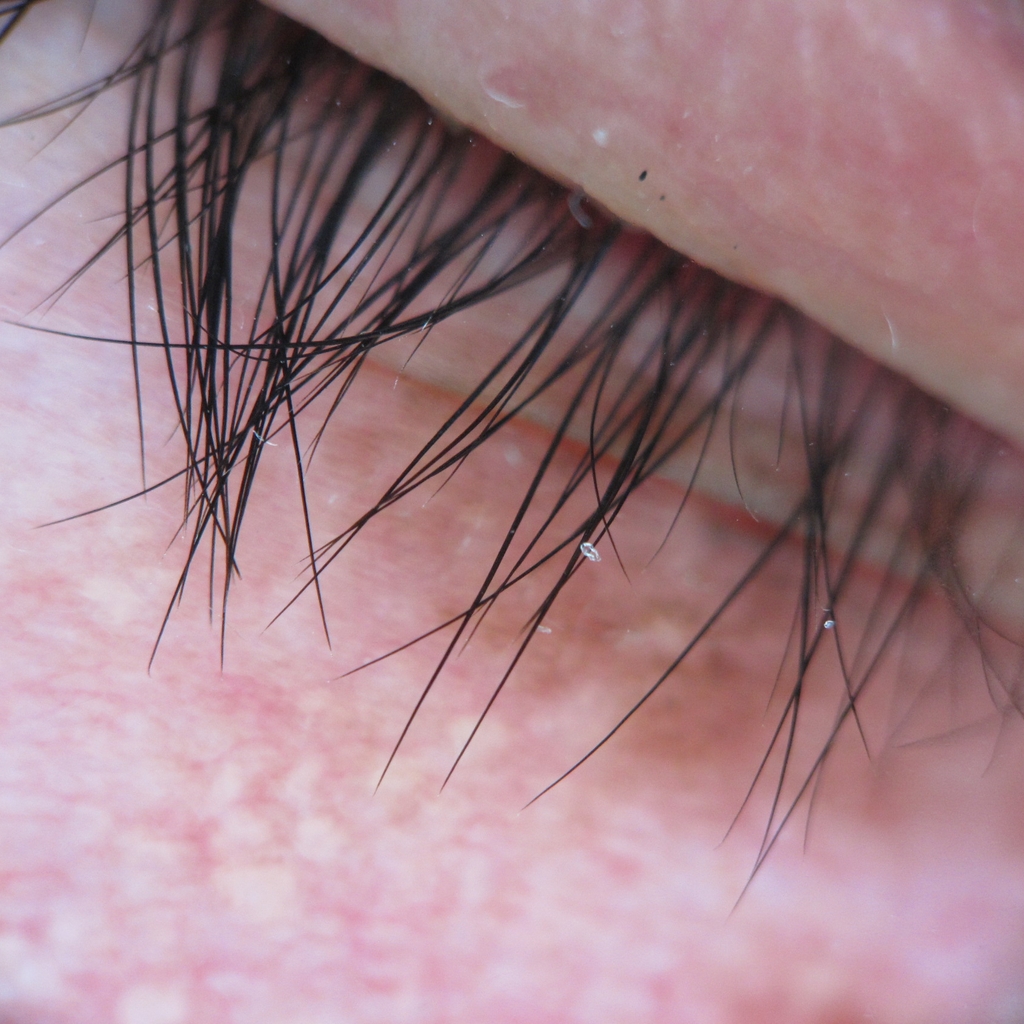}
        \caption{ISIC\_0058453}
        \label{fig:low_med_right}
    \end{subfigure}\hfill
    \begin{subfigure}{0.23\textwidth}
        \centering
        \includegraphics[width=\linewidth]{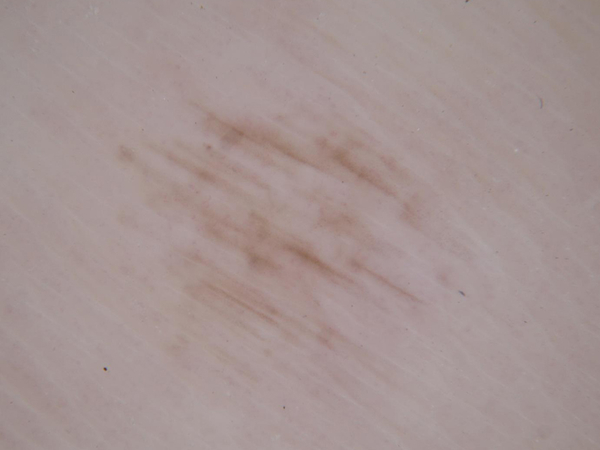}
        \caption{ISIC\_0027181}
        \label{fig:low_left}
    \end{subfigure}\hfill
    \begin{subfigure}{0.23\textwidth}
        \centering
        \includegraphics[width=\linewidth]{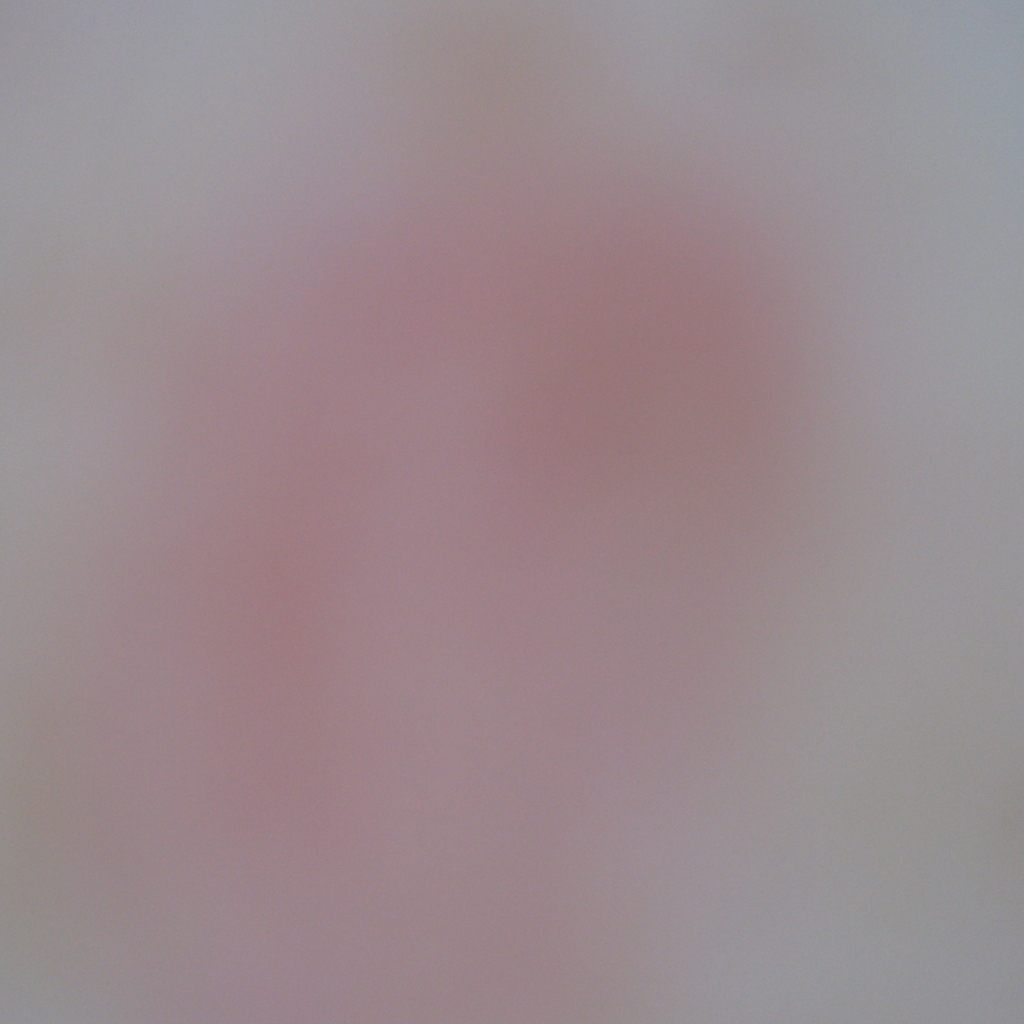}
        \caption{ISIC\_0067400}
        \label{fig:low_right}
    \end{subfigure}
    
    \caption{Examples of low-quality images excluded from the analysis. The first two images from the left (\subref{fig:low_med_left}, \subref{fig:low_med_right}) were considered to be of insufficient quality by expert dermatologists. The remaining two images (\subref{fig:low_left}, \subref{fig:low_right}) display blurred cases automatically identified by the blur score.}
    \label{fig:low_quality}
\end{figure}

Figure \ref{fig:low_quality} illustrates two representative samples from this collection. Typically, these images are characterized by blurriness or partial visibility of the pathological features. In response to the initial identification of these low-quality images, a systematic search for additional blurry images was conducted using analytical techniques such as the Fourier transform, Laplacian filter, and wavelet transform. 


Our analysis indicates that Laplacian- and wavelet-based measures are 
effective in distinguishing low-quality (blurred) dermatoscopic images 
from high-quality ones. In both cases, blurred images consistently 
receive lower scores compared to non-blurred images. This behavior
was not observed for the Fourier-based metric, whose values showed 
no clear separation between the two groups. To obtain a single, more
stable blur indicator, we standardized the outputs of the different 
transforms and combined them into a unified score.

\begin{figure}[]
    \centering
    \includegraphics[width=.5\linewidth]{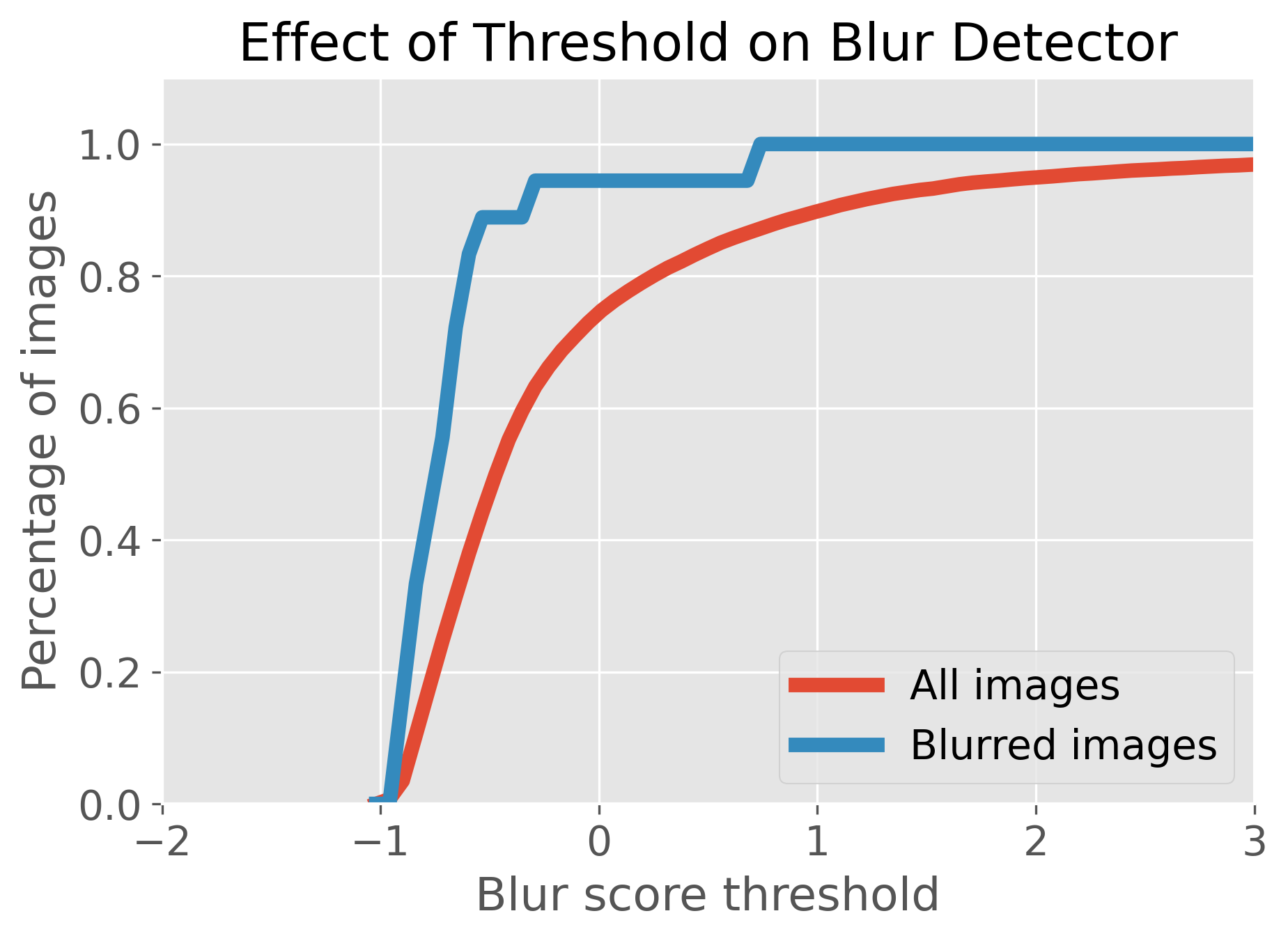}
    \caption{
    Effect of varying the blur-score threshold on the proportion of 
    images classified as blurry. The curves compare the cumulative 
    percentage of all images and dermatologist-annotated blurred 
    images as the threshold increases.
    }
    \label{fig:blur_threshold}
\end{figure}

Figure \ref{fig:blur_threshold} illustrates how varying the decision
threshold on this combined score affects the proportion of images 
classified as blurred. Even with relatively conservative thresholds, 
it is possible to discard most of dermatologist-labeled low-quality 
images while retaining the majority of high-quality ones. Although 
determining the optimal operating point requires further investigation, 
the visual inspection of the curves suggests that thresholds between 
–0.7 and –0.6 offer a reasonable trade-off. For instance, a threshold of 
–0.7 removes 61.11\% of blurred images while discarding only 25.90\% 
of high-quality ones, whereas a threshold of –0.6 increases the removal 
rate of blurred images to 83.33\% at the expense of excluding 36.82\% of 
high-quality images. We meticulously examined the 50 images exhibiting 
the lowest blurriness indices, as determined by a combination of methods,
we identify an additional 18 low-quality images,
two examples can be seen in Figure \ref{fig:low_quality}. 

The prevalence of low-quality images was notably higher among those categorized as difficult to classify, which could shed light on the frequent inaccuracies in model predictions for these images. The models often resort to "guessing" impeded by the inferior image quality. Nonetheless, a more systematic and automated approach for identifying low-quality images is imperative for future improvements. 


Although not highly effective for detecting image blur, the Fourier transform can be leveraged to identify images where hair occludes the skin lesion. Hair occlusion remains a significant challenge in dermoscopy; hair artifacts frequently obscure critical diagnostic information, such as lesion boundaries and textural patterns \cite{Toossi2013}. Consequently, the Fourier transform serves as a useful tool for isolating this specific subset of images. Furthermore, a visual inspection of these results reveals several duplicate images within the dataset. For instance, the first three images from the left in the top row (ISIC\_0056815, ISIC\_0066439, and ISIC\_0060839) clearly depict the same lesion, as do images ISIC\_0068642 and ISIC\_0061683, the first two in the second row.

\begin{figure}[]
    \centering
    \includegraphics[width=1\linewidth]{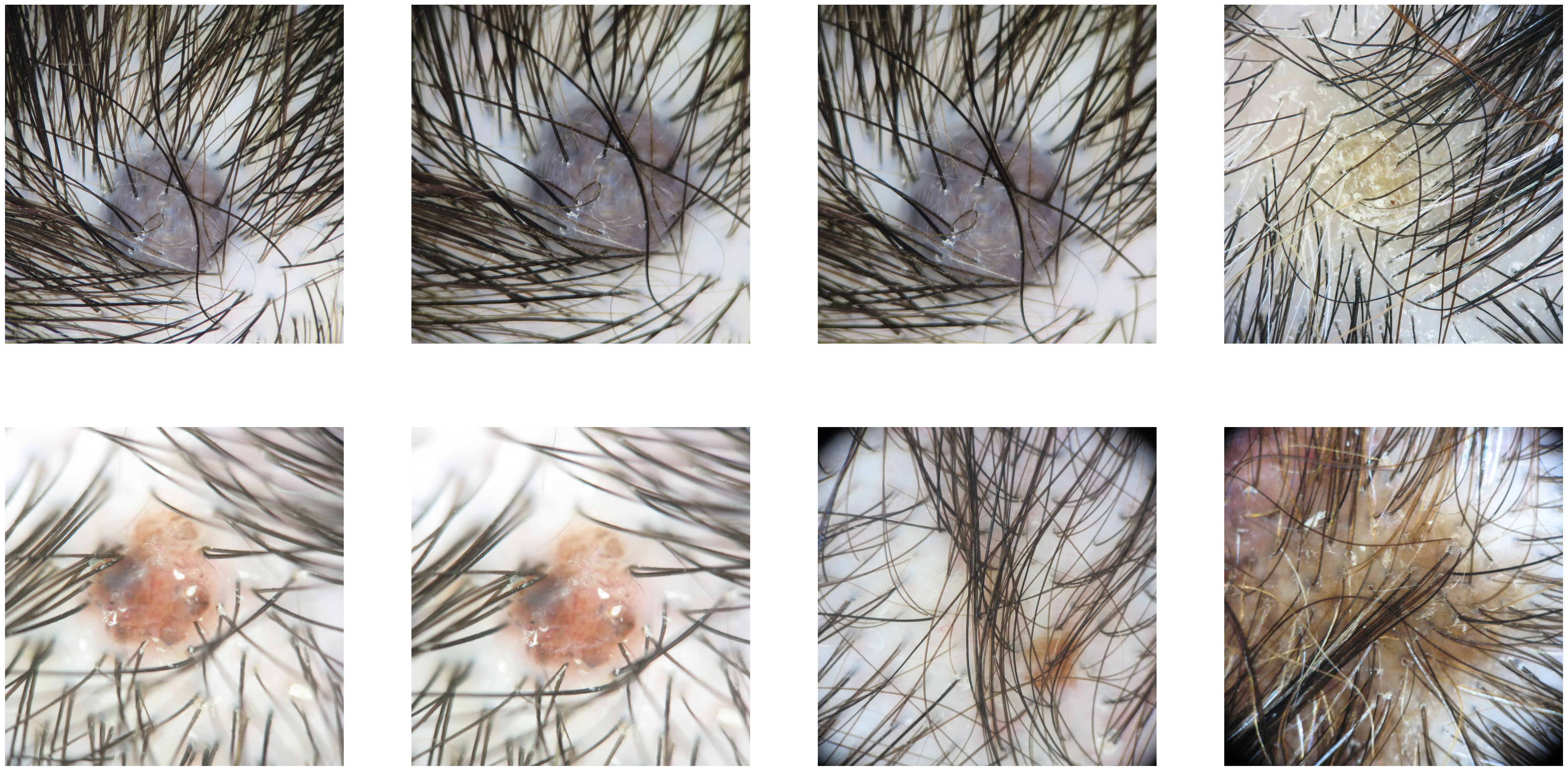}
    \caption{Images exhibiting strong high-frequency components, identified via Fourier transform. These images are heavily occluded by hair artifacts, and some instances are exact duplicates.}
    \label{fig:hairy}
\end{figure}

\section{Conclusion}
\label{conclusion}
In this study, we investigated dermatoscopic images that are consistently misclassified by both CNN models and expert dermatologists. Our results suggest that these cases reflect intrinsic diagnostic complexity, rather than merely limitations of the AI systems. A systematic analysis revealed that dermatologist label agreement on these difficult images was markedly lower than on the control set, with Cohen’s kappa dropping from 0.61 (control images) to 0.08 (difficult images). Inter-rater reliability followed the same trend: Fleiss’ kappa was 0.456 for the control images but decreased to 0.275 for the difficult subset, underscoring the inherent ambiguity and visual complexity of these cases. Moreover, when comparing the sensitivity and specificity reported in the literature, we observed that dermatologist performance on our control group aligns with or slightly exceeds previously published values, whereas performance declines substantially for the images that CNNs also misclassify, further confirming their intrinsic difficulty.

Image quality partially explained these classification challenges. The enhanced understanding and identification of these problematic cases could yield significant insights into diagnostic similarities between AI algorithms and human experts, other than further improvements in dermatology. The reason behind these common misclassifications remains an open question.

However, our study has several limitations, most notably the relatively small number of participating dermatologists.
This limited participation is partly due to the distinctive nature of our study design: unlike many previous works that ask clinicians to evaluate only a small subset of images, our protocol required each dermatologist to carefully assess 270 images and, in some cases, provide additional qualitative comments. This substantial workload inevitably restricted the number of experts we could involve. Future research should aim to address this limitation by incorporating larger and more diverse cohorts of dermatologists, thereby improving the robustness and external validity of the conclusions.

Another limitation concerns the identification of thresholds for detecting blurred images. Although we conducted a detailed analysis of image quality and duplicates, our assessment was not exhaustive, and further refinement of dataset curation strategies remains necessary. At present, blurred images can be identified either through manual inspection or by applying a threshold to the blur-detection scores. Future work could explore more sophisticated and fully automated approaches for identifying low-quality images, as well as empirically evaluate how improved image-quality filtering affects model performance on a more rigorously curated dataset.

Additionally, future work should actively involve dermatologists in the development of AI systems for skin disease classification to better simulate real-world clinical settings. We took preliminary steps in this direction by conducting an Exploratory Data Analysis (EDA) of patient information (Appendix \ref{a:eda_metadata}) and integrating these metadata into our ensemble learning framework. However, as reported in Appendix \ref{metaclassification}, this integration did not yield meaningful improvements in predictive accuracy. This outcome was unexpected from a clinical perspective, where demographic factors such as age, lesion location, and sex are known to strongly influence disease prevalence and diagnostic interpretation. A deeper investigation is required to clarify why these clinical variables failed to translate into measurable performance gains. Future research could leverage SHAP (SHapley Additive exPlanations) value analysis to elucidate which features drive the meta-classifier’s decisions and to quantify the true impact of patient metadata on the final predictions.


Furthermore, evaluating these algorithms in real-life clinical scenarios is critical to fully understand their practical performance. We believe in the necessity of rethinking comparative methodologies between human diagnosis and AI algorithms. To achieve a fair and relevant comparison, we propose evaluating AI algorithms on clinically assessed patients rather than solely relying on dermatoscopic images. Such an approach would introduce the challenge of OOD data and align the diagnostic context of AI with clinical realities, thereby enhancing the practical relevance of comparative analyses. Dermatologists’ diagnostic capabilities significantly benefit from in-person evaluations compared to image-only assessments \cite{Dinnes2018a}, incorporating additional contextual patient information and physical examination findings, such as palpation, is essential. 

Overall, this study highlights the importance of addressing intrinsic diagnostic challenges through improved dataset quality. We hope that our findings will help shape future research in several ways: (i) the identification of systematically difficult images may stimulate further investigation into the visual and clinical factors that make certain cases inherently ambiguous; and (ii) the observed link between image quality and diagnostic errors underscores the need for standardized and fully automated quality-assessment pipelines in real-world settings, as well as stricter data acquisition protocols.

\section*{Acknowledgment}
The authors thank Mr. Arturo Argentieri from CNR-ISASI Italy for his technical contribution to the GPU computing facilities. This research was funded in part by Future Artificial Intelligence Research—FAIR CUP B53C22003630006 grant number PE0000013.

\appendix

\section{Explanatory Data Analysis}
\label{a:eda_metadata}

\begin{figure}[]
    \centering
    
    \begin{subfigure}[b]{0.48\textwidth}
        \centering
        \includegraphics[width=\linewidth]{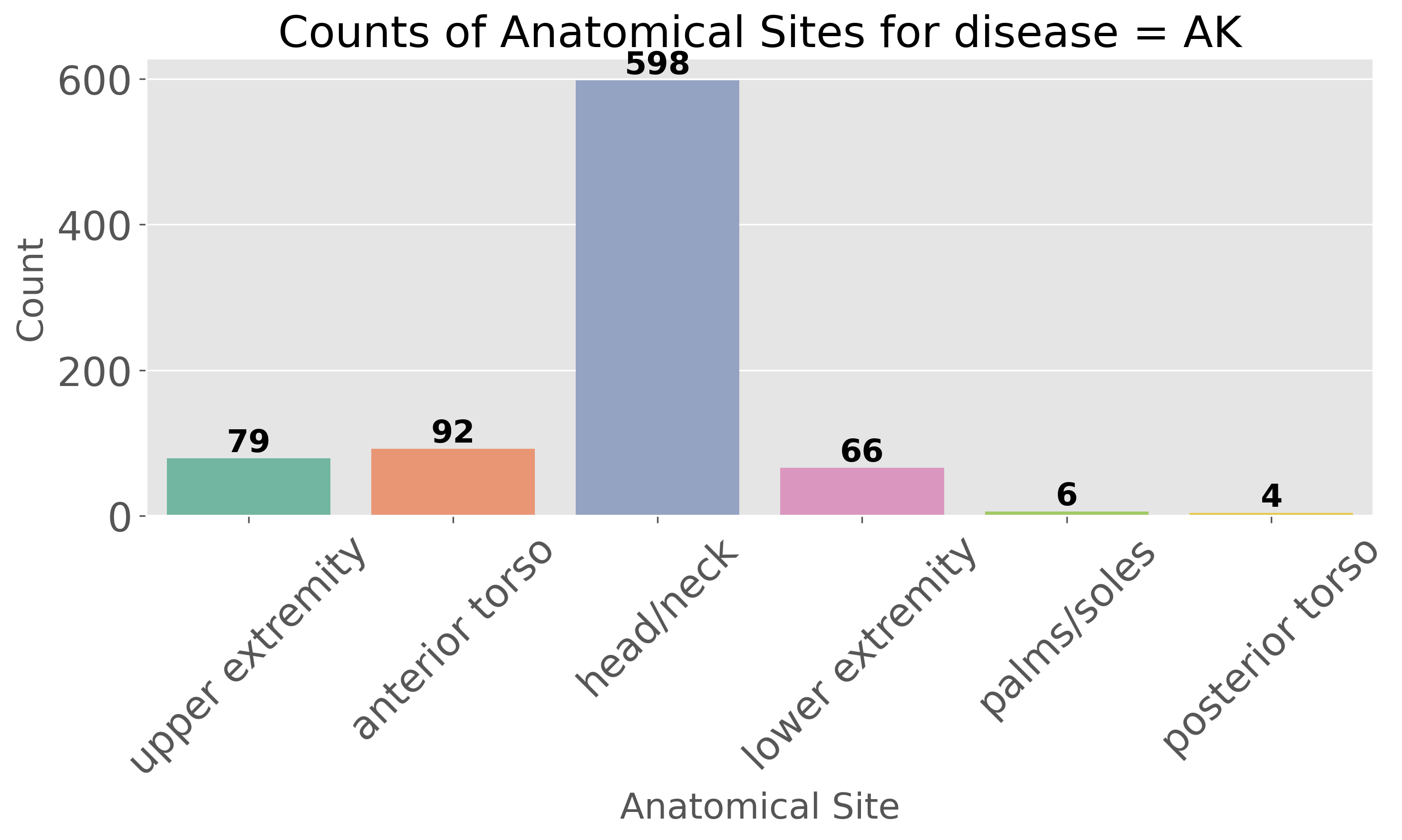}
        \caption{Frequency of Actinic Keratosis (AK) occurrences across dataset positions, with the x-axis representing the position index and the y-axis indicating the corresponding count of AK cases.}
        \label{fig_AK}
    \end{subfigure}
    \hfill 
    \begin{subfigure}[b]{0.48\textwidth}
        \centering
        \includegraphics[width=\linewidth]{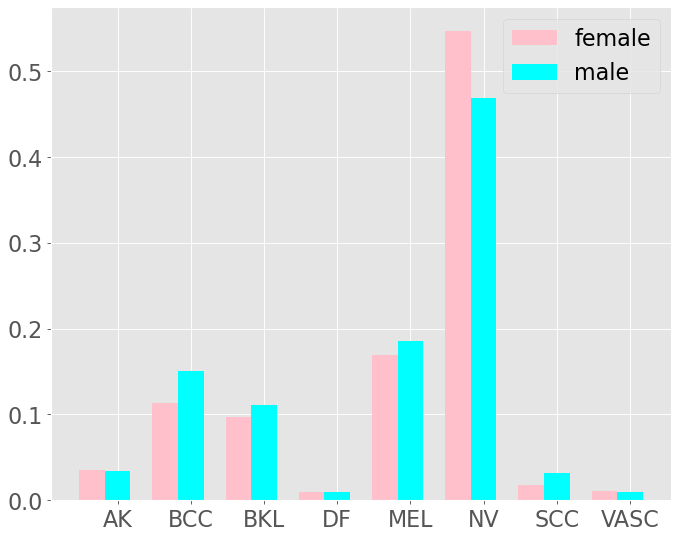}
        \caption{Comparison of the frequency of diseases normalized by gender. The count of occurrences of skin diseases by type, grouped by gender, has been divided by the total number of patients belonging to the same gender.}
        \label{fig_gender}
    \end{subfigure}
    
    \caption{Analysis of dataset distributions: (a) Actinic Keratosis frequency by position, and (b) normalized frequency of diseases by gender.}
    \label{fig_combined_dataset_analysis}
\end{figure}

An exploratory analysis of the dataset was conducted in collaboration with dermatologists to determine whether the inclusion of patient data could enhance model performance. Discussions with these medical professionals highlighted that certain skin diseases predominantly manifest on specific body parts and are more frequent in individuals of particular age groups and genders. This observation is corroborated by various studies in the existing literature \cite{Olsen2020, Fijakowska2021, Tognetti2021}. For example, as depicted in Figure \ref{fig_AK}, the majority of AK skin diseases were observed on the upper body. This contextual information could be leveraged by the algorithm to improve the accuracy of disease classification.

Similarly, age and gender also play critical roles in the prevalence and classification of skin diseases. According to Figure \ref{fig_gender}, benign diseases such as Nevus (NV) occur more frequently in women than in men. 

\begin{figure}[]
\centering
\begin{subfigure}{.5\columnwidth}
  \centering
  \includegraphics[width=\linewidth]{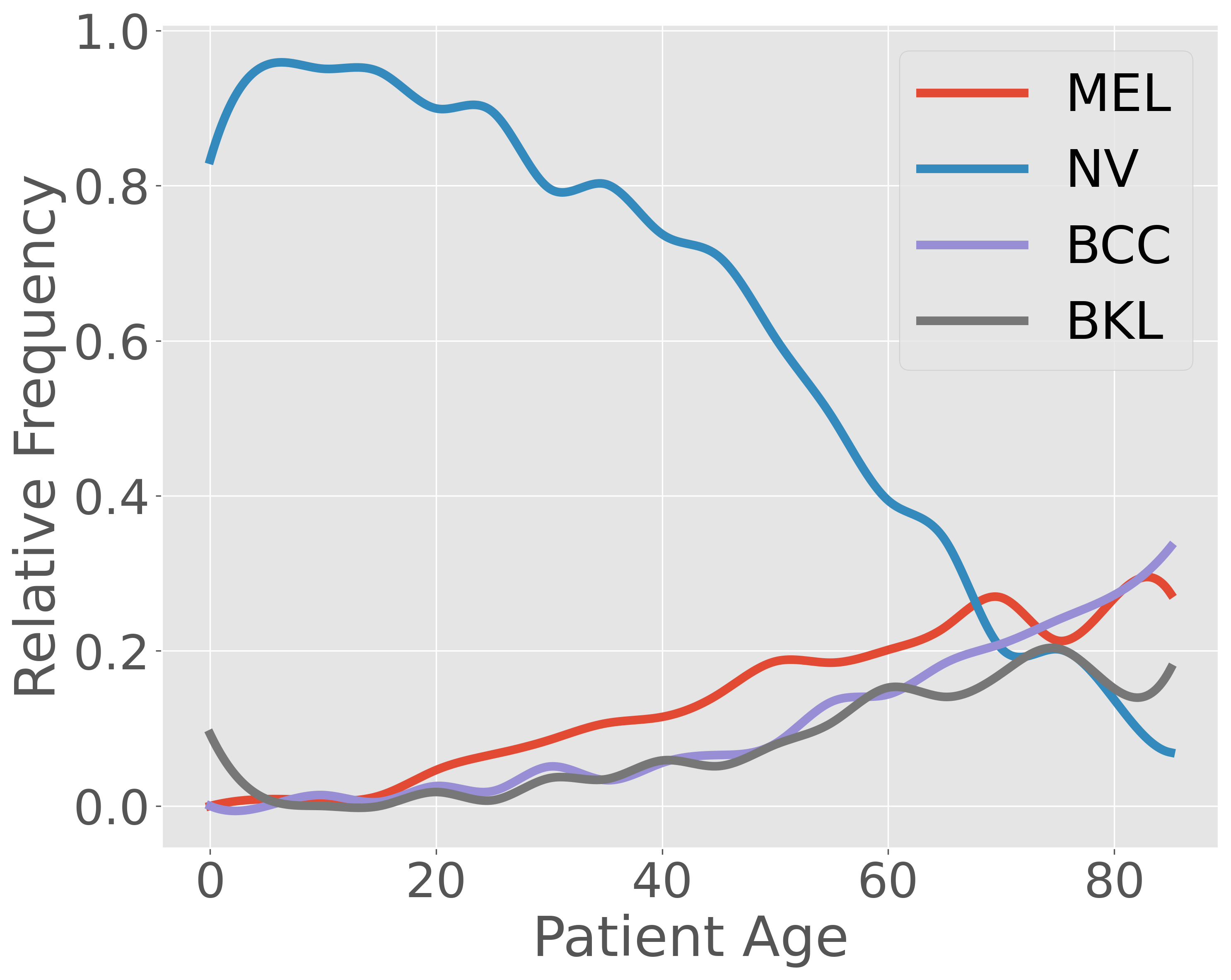}
  \caption{Most common disease in the dataset.}
  \label{fig:age_common}
\end{subfigure}%
\begin{subfigure}{.5\columnwidth}
  \centering
  \includegraphics[width=\linewidth]{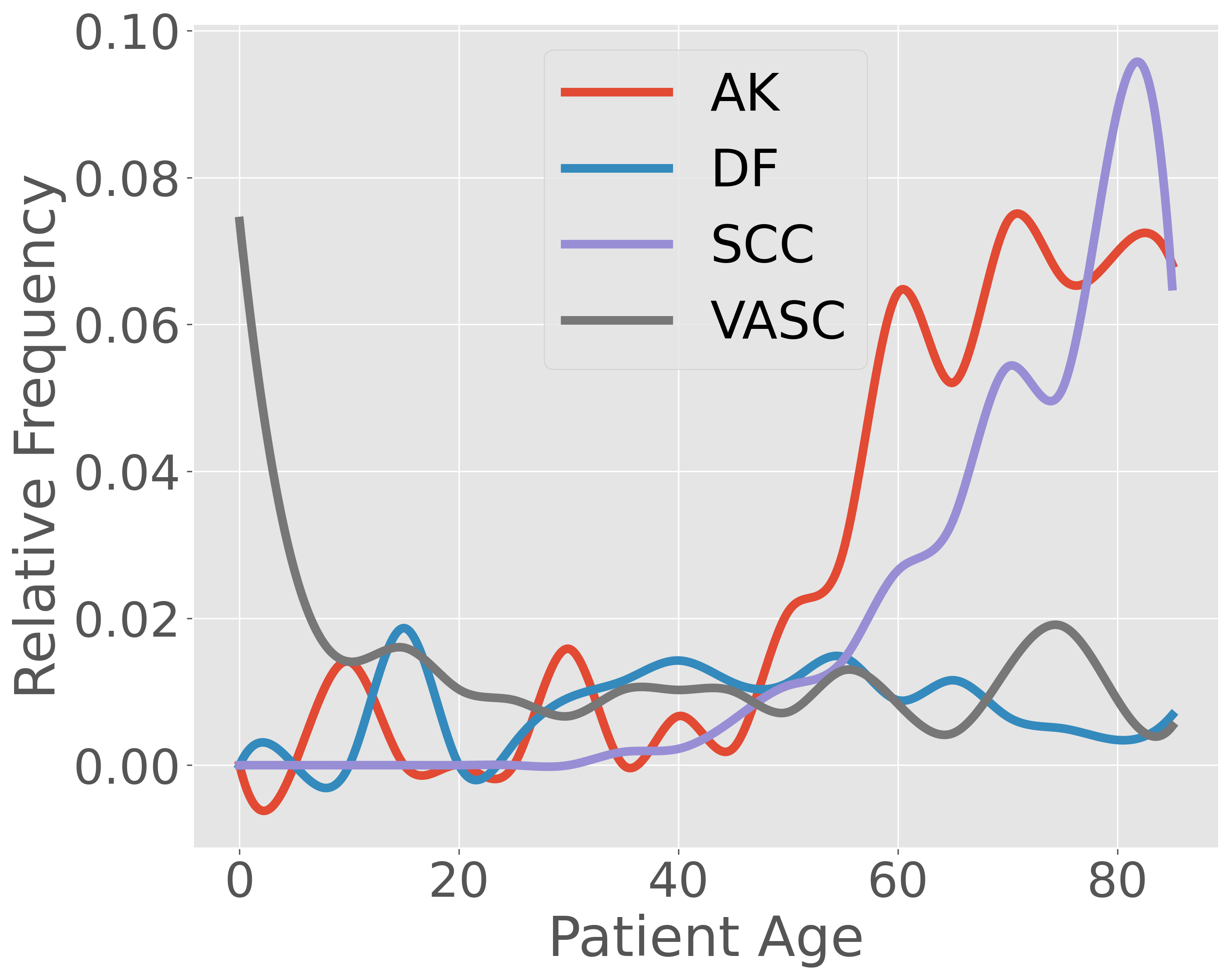}
  \caption{Less frequent disease in the dataset.}
  \label{fig:age_rare}
\end{subfigure}
\caption{Frequency of diseases normalized by age. The number of patients for each skin disease grouped by age was divided by the total number of patients for the corresponding age. The curve are smoothed to improve visualization.}
\label{fig:age}
\end{figure}

Additionally, Figure \ref{fig:age} demonstrates that up to the age of 40, the majority of skin conditions are benign. However, this trend shifts after the age of 70, where the incidence of malignant diseases increases significantly. These demographic insights could be instrumental in refining the predictive capabilities of diagnostic algorithms. More extensive studies on data can be found in our GitHub repository.

\section{Website}
\label{a:website}

This appendix presents screenshots of the custom web application developed to facilitate the clinical evaluation. To ensure complete flexibility in their assessments, participating dermatologists were able to submit, review, and modify their diagnoses at any time. Additionally, the interface allowed users to view the dermoscopic images at full resolution for a more detailed inspection when needed.

\begin{figure}[]
\centering
\begin{subfigure}{.40\columnwidth}
  \centering
  \includegraphics[width=\linewidth]{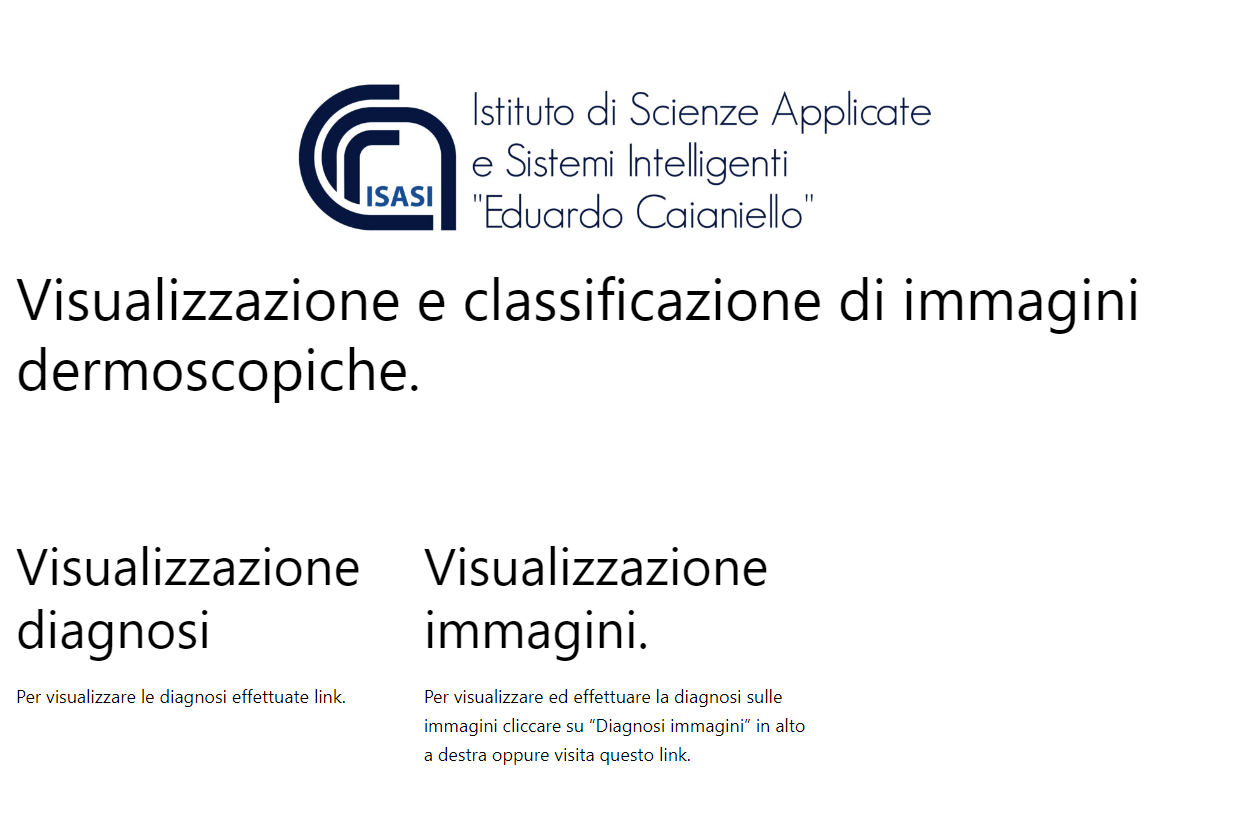}
  \caption{Landing page of the diagnostic platform.}
  \label{fig:sub1}
\end{subfigure}%
\begin{subfigure}{.40\columnwidth}
  \centering
  \includegraphics[width=\linewidth]{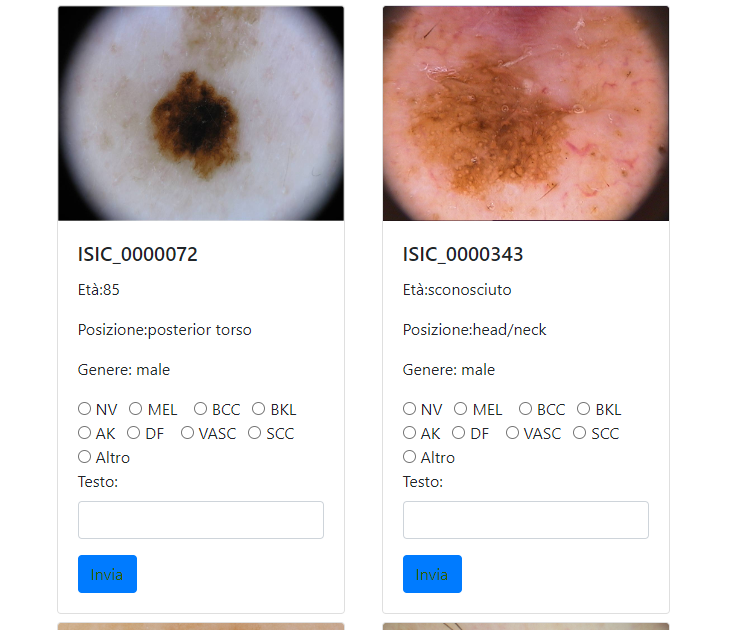}
  \caption{Diagnosis interface allowing to review patient images and patient data.}
  \label{fig:sub2}
\end{subfigure}
\caption{Screenshots of the web-based diagnostic platform developed for this study, designed to facilitate dermatologists’ assessments of challenging images while maintaining data security and user-friendliness.}
\label{fig:test}
\end{figure}

\section{Ensemble and Meta-classification}
\label{metaclassification}

More than the assessment of \textit{difficult images}, we conduct a detailed investigation on how to improve the AI performance considering several key elements: using individual model, ensemble techniques, meta-classifiers, and the integrated performance of meta-classifiers when combined with patient data. This appendix aims to highlight the relative advantages and performance metrics of each approach, offering a comprehensive evaluation of their effectiveness in classification tasks. We the compare our results with the state-of-art algorithms.

After the training phase of single models, a comprehensive dataset was compiled using the respective predictions on the validation sets. Specifically, these predictions refer to the class activations produced by each model, defined as the outputs from the fully connected layer immediately preceding the application of the softmax function.

Using this dataset, we investigated the utility of ensemble methods, which are known to enhance generalization performance and broadly adopted in the literature. This ensemble methodology offered dual advantages: firstly, it ensured comprehensive coverage, as all images from the initial dataset were utilized for model training; secondly, it guaranteed partial distinctiveness of training data across models, further enhancing generalizability.

Additionally, a meta-classifier was employed to mitigate biases inherent in individual models and to incorporate patient data information without retraining all the models. The derived predictions served as input features for training several classifiers, including Decision Tree \cite{Quinlan1986}, Random Forest (RF)\cite{Breiman2001}, eXtreme Gradient Boosting (XGB) \cite{Chen2016}, CatBoost \cite{catboost}, and Multilayer Perceptron (MLP).


In subsequent experiments, patient data—specifically gender, age, and disease location—were integrated into the dataset. Missing patient data were imputed using the mode of the corresponding feature within each disease class. To encode categorical variables effectively OneHotEncoding was employed.


We found the best meta-classifiers' hyperparameters using a 10-fold cross-validation strategy. Additionally, 20\% of the dataset was reserved exclusively as an external test set, thus remaining entirely independent of the cross-validation process. To further enhance the robustness and stability of the evaluation, the entire cross-validation procedure was repeated using five different random seeds.

To statistically evaluate the impact of patients' data and the 
performance differences among the various models. The analysis is 
based on the Friedman test, which assesses whether performance 
disparities among models are unlikely to have arisen by chance. 
When the Friedman test indicates significant overall differences, 
the Nemenyi post-hoc test is applied to conduct pairwise 
comparisons and determine which model pairs differ significantly 
in their average ranks.


\begin{figure}[]
\centering
\includegraphics[width=0.9\textwidth]{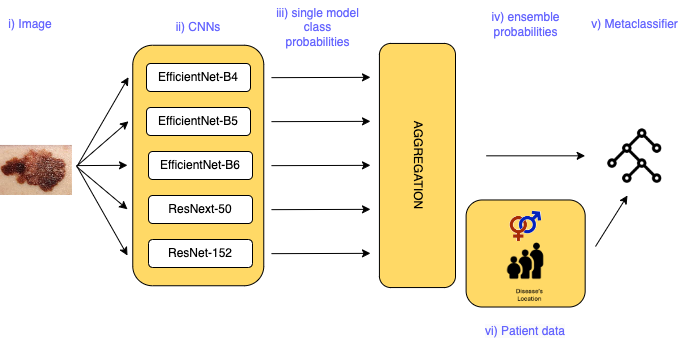}
\caption{The architecture utilized during the training phase is schematically represented in the diagram. Initially, the input image (i) is fed into various convolutional neural networks (ii). Each CNN (Convolutional Neural Network) processes the image and outputs probabilities/activations for each class (iii). Subsequently, in step (iv), an aggregation method is applied to these probabilities to generate an ensemble prediction. Finally, in step (v), a meta-classifier is trained using the aggregated probabilities from the ensemble. Additionally, (vi) patient data is incorporated.}
\label{fig_training_architecture}
\end{figure}

The overall model architecture employed during training is illustrated in Figure~\ref{fig_training_architecture}, highlighting its key components and structural details to enhance clarity and understanding of the proposed system. At inference time, the architecture mirrors the training configuration, with one critical additional step. Before processing an image through the CNNs (step ii), predictions from models sharing the same architecture but trained on distinct cross-validation folds are averaged. 

\subsection{Classification performance}
\label{subsec:classification}
In this section we will present the results of the the models, comparing the various techniques each other and with recent papers in the literature. The categorization of the model results is based on the architectural types used. We designate models as 'Single Models' when predictions are generated using a single CNN, as illustrated up to step (iii) in Figure \ref{fig_training_architecture}. 'Ensemble' models integrate the outputs of multiple CNNs via a non-machine learning technique, continuing up to step (iv). 'Meta-Classifier' models employ an additional model trained on the CNNs’ class activation, as shown up to step (v) in the architecture diagram. Finally, 'Meta-Classifier + Patient Data' models incorporate both CNN predictions and patient data, extending to step (vi) in Figure \ref{fig_training_architecture}. 

Table \ref{table1} summarizes the performance of various neural networks evaluated on test set. 
We utilized balanced accuracy as it was the primary evaluation metric employed in the ISIC Challenge leaderboard. 

\begin{table}[]
\centering
\caption{
Balanced accuracy (BA) values (mean ± standard deviation) for all evaluated models, including single CNN architectures, ensemble methods—where MP denotes mean prediction and MV denotes majority voting—and meta-classifiers (MC), and meta-classifiers with patient data (MC + PD). MLP refers to a multilayer perceptron. In bold the results with the higher mean.
}
\label{table1}
\resizebox{0.5\textwidth}{!}{
\begin{tabular}{l|l|c}
\hline
\textbf{Group} & \textbf{Model} & \textbf{BA (\%)} \\
\hline
\multirow{5}{*}{\textbf{Single}} 
    & EfficientNetB4 & $85.36 \pm 1.98$ \\
    & EfficientNetB5 & $84.91 \pm 1.71$ \\
    & EfficientNetB6 & $83.42 \pm 0.97$ \\
    & ResNext50      & $80.80 \pm 0.88$ \\
    & ResNet152      & $79.05 \pm 1.99$ \\
\hline
\multirow{2}{*}{\textbf{Ensemble}} 
    & MP  & $87.53$ \\
    & MV   & $86.90$ \\
\hline                                
\multirow{5}{*}{\textbf{MC}} 
    & Decision Tree  & $85.67 \pm 0.79$ \\
    & \textbf{Random Forest}  & $\mathbf{89.42 \pm 0.75}$ \\
    & XGBoost        & $87.55 \pm 0.69$ \\
    & CatBoost       & $86.78 \pm 0.91$ \\
    & MLP            & $86.19 \pm 1.02 $\\                      
\hline
\multirow{5}{*}{\textbf{MC + PD}} 
    & Decision Tree   & $85.57 \pm 0.66 $ \\
    & Random Forest   & $ 89.25\pm 0.52 $ \\
    & XGBoost         & $ 87.37 \pm 0.63$\\
    & CatBoost        & $86.89 \pm 0.80 $ \\  
    & MLP             & $86.94 \pm 0.64$ \\                          
\hline
\end{tabular}
}
\end{table}

The findings demonstrate that integrating multiple CNNs into the predictive framework significantly improves classification accuracy. Individual CNN models yielded performance metrics ranging between 79.05\% and 85.36\%; however, the ensemble approach notably increased balanced accuracy to 87.53\%. This enhancement is particularly evident in the Mean Prediction (MP) case.

Moreover, the use of a meta-classifier to aggregate the predictions 
produced by the ensemble leads to additional performance improvements
when employing a Random Forest as the final classifier, reaching a 
balanced accuracy of 89.75\%. For other meta-classification models,
however, the benefit is less pronounced, particularly in the case of 
the Decision Tree. This pattern offers insight into the underlying 
bias–variance dynamics: Random Forests, through bootstrapping and
ensemble averaging, exhibit relatively low bias and reduced variance, 
whereas single Decision Trees are highly sensitive to data fluctuations 
and thus prone to high variance. Given that Random Forest is 
the best-performing meta-classifier and the Decision Tree the 
least effective, it is reasonable to speculate that the prediction
space generated by the models forms a dataset with inherently high
variance, which favors ensemble-based learners over single-tree methods.

Several reasons could explain these findings. Firstly, ensemble methods effectively reduce individual model biases and errors by combining diverse perspectives, thereby enhancing robustness and generalization capabilities. Furthermore, meta-classifiers are capable of identifying complex patterns and interactions among predictions from different models, providing a more nuanced integration of model outputs. 

\begin{figure}[]
    \centering
    \includegraphics[width=.5\linewidth]{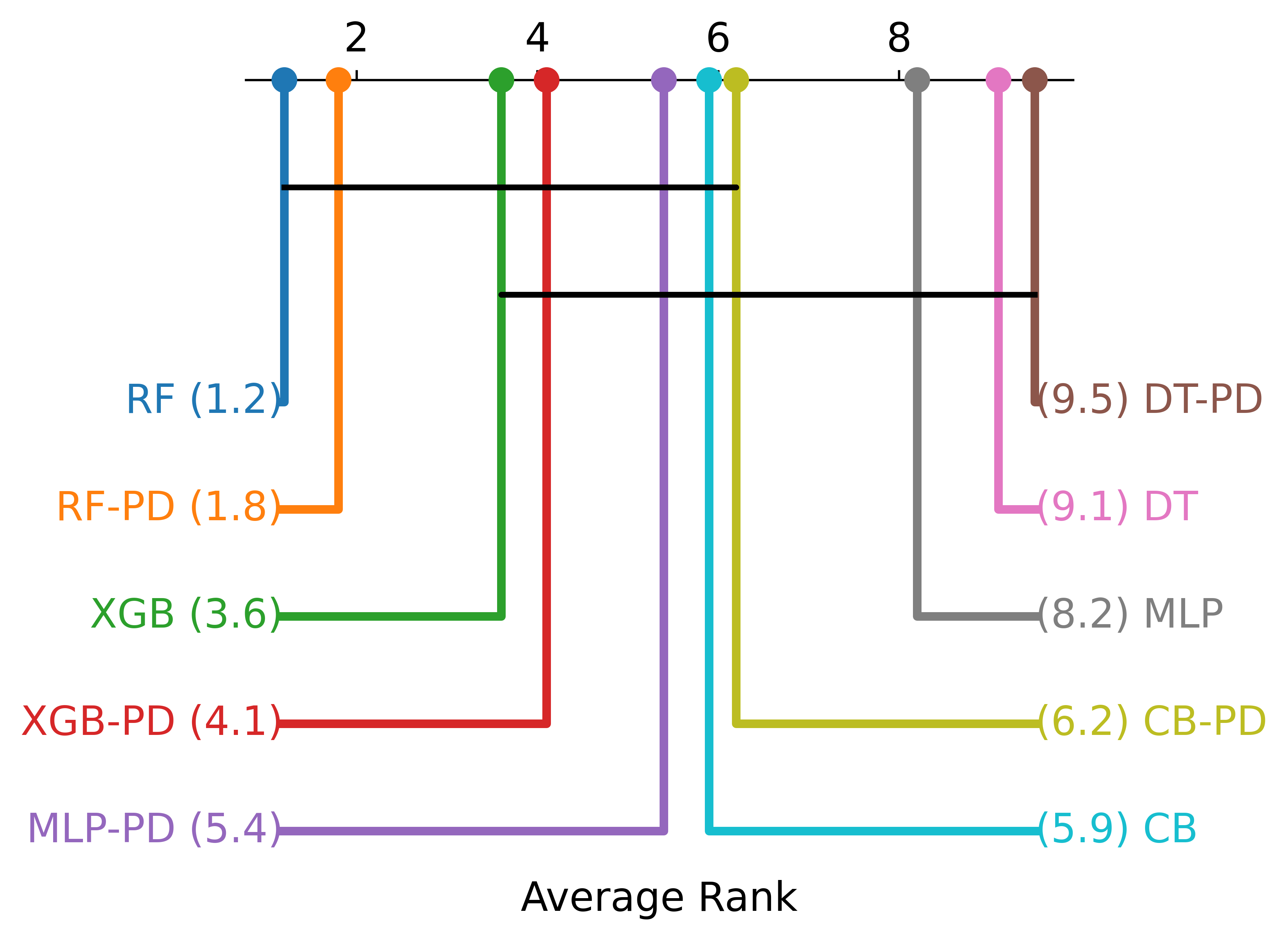}
    \caption{
    Critical Difference (CD) diagram reporting the average ranks of the evaluated models across all experimental datasets. Lower ranks indicate better predictive performance. Horizontal connectors identify groups of models whose performance differences are not statistically significant according to the Nemenyi post-hoc test at the 5\% significance level. "PD" represent the model that involve patient data.
    }
    \label{fig:statistical_ranking}
\end{figure}

We employed Critical Difference (CD) plot to have a more statistical comparison of the models, we exclude single model for the evident difference in performance and we exclude the simple ensemble methods because it was possible only to have a single test results. We can observe that between the best model RF and the sixth model there are no statistical difference.

The CD plot highlights that the incorporation of patient data into the predictive model do not significally increases classification accuracy. This unexpected results because patient information is used in clinical practice.




\begin{table}[]
\caption{
Performance comparison of various methods evaluated on the ISIC 2019 dataset in terms of F1-Score Macro, Precision Macro, and Recall Macro. Results marked with (*) are sourced from the benchmark study conducted by Ghazouani et al. \cite{Ghazouani2025}.
}
\label{tab:comparison}
\centering
\resizebox{0.5\textwidth}{!}{
\begin{tabular}{lccc}
\hline
\textbf{Method} & \textbf{F1-Score} & \textbf{Precision} & \textbf{Recall}\\ 
\hline
ResAttentionNet* \cite{Wang2017} & 0.82 & - & - \\
ARL-CNN* \cite{Zhang2019} & 0.80 & - & - \\
Soft* Attention CNN  \cite{Alhudhaif2023}& 0.77 & - & - \\
IARO* \cite{AbdElaziz2023} & 0.78 & - & - \\
S$^2$C-DeLeNet* \cite{Alam2022} & 0.81 & - & - \\
X-R50* \cite{Panthakkan2022} & 0.81 & - & - \\
MuRANet* \cite{Ghazouani2025} & 0.88 & - & - \\ 
Ozdemir et. al\cite{Ozdemir2025} & \textbf{0.9182} & \textbf{0.9324} & \textbf{0.9070} \\
\hline
Our (MLP) & 0.8834 & 0.8809 & 0.8869 \\
Our (RF) & 0.8611 & 0.8295 & 0.9003 \\
\end{tabular}
}
\end{table}

Table~\ref{tab:comparison} presents a comparative analysis of various state-of-the-art methodologies with respect to their ROC-AUC and F1-score performance metrics. The proposed methods, an MLP and a RF, achieve superior ROC-AUC scores of 0.99, outperforming all benchmarked approaches that reports this metric. Notably, MuRANet demonstrates the highest performance among previously reported methods, with a ROC-AUC of 0.95 and an F1-score of 0.88. However, the method recently proposed by Ozdemir et al. \cite{Ozdemir2025} attains the highest recorded F1-score (0.9182), precision (0.9324), and recall (0.9070). In the Table are reported the RF and MLP meta-classifiers results using also patient information.

\begin{figure}[]
\centering
\includegraphics[width=.5\linewidth]{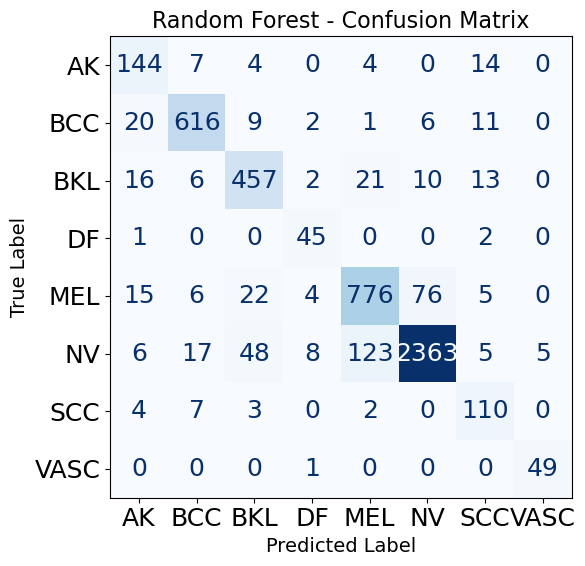}
\caption{Confusion matrix of the Random Forest model.}
\label{fig:cm_rf}
\end{figure}

Figure~\ref{fig:cm_rf} displays the confusion matrix from the RF. RF model is characterized by superior balanced accuracy. Despite this advantage, the relatively lower F1-score suggests a potential trade-off of MLP.

It is also noteworthy that all the aforementioned methods are orthogonal to our approach. Consequently, they can readily be integrated as individual classifiers within our ensemble framework, replacing or complementing the general-purpose CNN currently employed. This integration would enable leveraging advanced convolutional and attention-based layers for extracting high-level image features, while simultaneously exploiting the benefits of ensemble learning and incorporating patient-specific data.





\clearpage 




\printcredits

\bibliographystyle{cas-model2-names}

\bibliography{biblio}



\end{document}